\documentclass[letterpaper, 10 pt, conference]{ieeeconf}
\IEEEoverridecommandlockouts
\usepackage{amsmath} 
\usepackage{amssymb}  
\usepackage{float}
\usepackage{graphicx}
\usepackage{algorithm,algorithmic}

\title{\LARGE TEAM: Trilateration for Exploration and Mapping with Robotic Networks}

\author{Lillian Clark$^{1}$, Charles Andre$^{1}$, Joseph Galante$^{2}$, Bhaskar Krishnamachari$^{1}$, and Konstantinos Psounis$^{1}$
\thanks{This work was supported by a NASA Space Technology Research Fellowship Grant No. 80NSSC19K1189.}
\thanks{$^{1}$Authors are with the Ming Hsieh Department of Electrical and Computer Engineering, University of
Southern California.
        {\tt\small \{lilliamc, candre, bkrishna, kpsounis\}@usc.edu}}%
\thanks{$^{2}$Author is with NASA Goddard Space Flight Center.
        {\tt\small joseph.m.galante@nasa.gov}}%
}
\begin{document}
\maketitle
\thispagestyle{empty}
\pagestyle{empty}

\begin{abstract}
    Motivated by lunar exploration, we consider deploying a network of mobile robots to explore an unknown environment while acting as a cooperative positioning system. Robots measure and communicate position-related data in order to perform localization in the absence of infrastructure-based solutions (e.g. stationary beacons or GPS).
    We present Trilateration for Exploration and Mapping (TEAM), a novel algorithm for low-complexity localization and mapping with robotic networks. TEAM is designed to leverage the capability of commercially-available ultra-wideband (UWB) radios on board the robots to provide range estimates with centimeter accuracy and perform anchorless localization in a shared, stationary frame. It is well-suited for feature-deprived environments, where feature-based localization approaches suffer.
    We provide experimental results in varied Gazebo simulation environments as well as on a testbed of Turtlebot3 Burgers with Pozyx UWB radios. We compare TEAM to the popular Rao-Blackwellized Particle Filter for Simultaneous Localization and Mapping (SLAM). 
    We demonstrate that TEAM requires an order of magnitude less computational complexity and reduces the necessary sample rate of LiDAR measurements by an order of magnitude. These advantages do not require sacrificing performance, as TEAM reduces the maximum localization error by 50\% and achieves up to a 28\% increase in map accuracy in feature-deprived environments and comparable map accuracy in other settings. 
\end{abstract}

\section{Introduction}
Scientists believe tunnels formed by cooled flowing lava exist below the surface of the moon and may be a favorable environment for human activities \cite{horz1985lava}.
Multi-robot systems have been proposed to explore these harsh, remote environments because of the low volume, mass, and development costs of small robots and the inherent redundancy of teams of robots \cite{husain2013mapping, whittaker2012technologies, vaquero2018approach}.
Cooperation in robotic networks presents further advantages in three dimensions: networking (robots can forward data to extend the effective communication range), positioning (robots can collect inter-robot range measurements), and task performance (multiple robots can improve mapping efficiency) \cite{yan2013survey, leitner2009multi, olfati2007consensus,kumar2016networked, kurazume1994cooperative}.

\begin{figure}
    \centering
    \includegraphics[width=0.8\columnwidth]{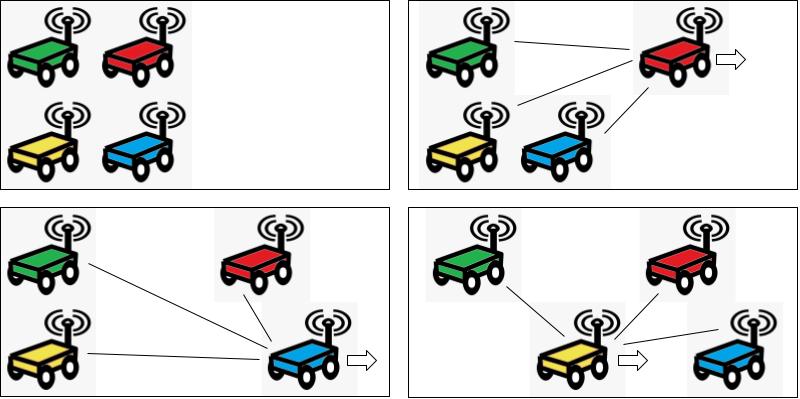}
    \caption{TEAM leverages collaborative localization and coordinated mobility. Top left: robots initiate a shared coordinate frame. Top right: the first robot moves and maps while using its neighbors to trilaterate. Bottom left and right: subsequent robots take turns moving and mapping.}
    \label{fig:overview_diagram}
\end{figure}

We focus on the application of mapping unknown environments with light detection and ranging (LiDAR) data and transferring this data to a stationary data sink (e.g. a lunar lander). 
We use ultra-wideband (UWB) radios configured to perform two-way ranging (TWR), i.e. exchanging packets and measuring round-trip time to estimate distance \cite{dewberry2015precision}.
A popular solution for localization in GPS-denied environments (e.g. indoors), UWB positioning allows a robot to determine its position relative to the known positions of three other UWB radios, called anchors or beacons \cite{gezici2005localization, alarifi2016ultra, nguyen2016ultra}. 

Lunar exploration presents additional challenges beyond the lack of GPS in that the establishment of UWB anchor infrastructure or other stationary positioning references is infeasible \cite{whittaker2012technologies}. However, in this setting each mobile robot can act as an instantaneous anchor for cooperative positioning (Fig. \ref{fig:overview_diagram}). This approach is also applicable for settings where GPS information is unavailable or insufficient such as precision agriculture, operating in dense urban environments, and search and rescue \cite{nguyen2016ultra}.

We present Trilateration for Exploration and Mapping (TEAM), a novel approach to localization and mapping which leverages the advantages of robotic networks. Via coordinated mobility, ad hoc multi-hop communication, and the ability to trilaterate a position estimate, we demonstrate that four or more resource-constrained robots can develop accurate maps of unknown environments. Our contribution is the design and experimental evaluation of TEAM which (1) requires an order of magnitude less computational complexity than a Rao-Blackwellized Particle Filter \cite{grisetti2007improved, gmapping}; (2) reduces the necessary sample rate of LiDAR measurements by an order of magnitude; (3) can reduce the maximum localization error by 50\%; and (4) achieves up to a 28\% increase in map accuracy in feature-deprived environments and comparable map accuracy in other settings. These computational, power consumption, and performance advantages are obtained at the cost of overall task efficiency, as the robots are periodically static in order to trilaterate. We demonstrate TEAM in several simulation environments and evaluate its performance on a network of mobile robots.

\section{Related Work}
Localization and mapping in multi-robot systems is a well-explored research area \cite{saeedi2016multiple}.
Previous work seeks to improve accuracy of localization and the resulting maps through collaboration, wherein robots communicate their maps, positions, or other relevant data \cite{kia2016cooperative,prorok2014accurate}.
Inter-robot detection and/or ranging can be used to improve localization \cite{rekleitis2001multi,prorok2014accurate}. 
While previous work has sought to augment state estimation algorithms with inter-robot ranging, focusing on improved accuracy \cite{wang2017ultra,hoeller2017augmenting}, 
in this work we focus on low-complexity approaches and examine the feasibility of inter-robot ranging as the primary method for absolute localization.

Recently, UWB has become a popular solution for localization in GPS-denied environments like factories and warehouses \cite{gezici2005localization,alarifi2016ultra}. 
One significant advantage of UWB is that it offers the ability for low-complexity trilateration \cite{zhou2009efficient}.
Shule \textit{et al.} presented a survey of UWB localization for collaborative multi-robot systems and highlighted recent trends \cite{shule2020uwb}. 
Most existing work relies on a set of fixed anchors with known positions, and uses Kalman filters or least squares estimators for tracking robots \cite{nguyen2016ultra}.
Fusing UWB ranging data from these stationary anchors with other sensors is a popular research direction \cite{shule2020uwb,wang2017ultra}. 

Several prior works consider self-localization in UWB anchor networks. Di Franco \textit{et al.} presented a method for calibration-free, infrastructure-free localization in sensor networks based on inter-node UWB ranging \cite{di2017calibration}. 
Hamer and D'Andrea presented a self-localizing network of UWB anchors which then allow multiple robots to localize based on received signals \cite{hamer2018self}. 
Subramanian and Lim presented a scalable distributed localization scheme and introduce the possibility of anchor node mobility, tested in simulation \cite{subramanian2005scalable}. 
Our work extends this idea to the application of mapping and evaluates it experimentally.

While stationary UWB anchors have received considerable attention for robotics in recent years, few works consider UWB transceivers onboard the robots for relative positioning. As a method for inter-robot ranging, UWB has recently been demonstrated for autonomous docking \cite{nguyen2019integrated}, 
formation flying \cite{guo2019ultra}, 
and leader-follower \cite{nguyen2019distance,cao2018uwb}.
In a closely related work to ours, Guler \textit{et al.} experimentally evaluate the use of three UWB transceivers on a single robot and one UWB transceiver on a second robot to perform accurate relative localization without explicit inter-robot communication \cite{guler2019infrastructure}. 
In another closely related work, Kurazume and Hirose propose an approach to cooperative positioning in which a group of robots is divided into two groups who alternate acting as reference landmarks, which they demonstrate on three robots using a laser rangefinder for inter-robot ranging \cite{kurazume2000experimental}. Inter-robot detection in this manner introduces the correspondence problem of matching the detected robot with its identifier, which is difficult to address for large swarms. Using UWB with unique device addresses solves this issue.
Our work combines the accurate relative localization of UWB demonstrated by Guler \textit{et al.} \cite{guler2019infrastructure} and coordinated mobility for cooperative positioning demonstrated by Kurazume and Hirose \cite{kurazume2000experimental}. Furthermore, we introduce explicit inter-robot communication and consider the application of mapping unknown environments.

\section{Trilateration for Exploration and Mapping}


TEAM is described in Algorithm \ref{alg:team}. After initializing position estimates, each robot iterates through lines 2-12 wherein they map their environment and exchange their coordinates and maps. During a robot's TDMA window (see Fig. \ref{fig:TDMAwindsows}), the robot drives autonomously, initiates two-way ranging, and trilaterates. The following subsections present each component of the algorithm.

\begin{algorithm}
 \caption{TEAM running on robot $i$}
 \label{alg:team}
 \begin{algorithmic}[1]
   \STATE coords, neighborCoords = Initialization()
   \WHILE {True}
       \STATE neighborCoords = RadioReceive()
       \IF {currentTime is in $\textrm{TDMAwindow}_i$}
       \STATE odom = DriveAutonomously()
       \STATE neighborDistances = UWBRange()
       \STATE coords = TrilateratePosition(odom, neighborCoords, neighborDistances)
       \ENDIF
       \STATE scan = LiDARScan()
       \STATE map = UpdateMap(coords, scan, map)
       \STATE RadioTransmit(coords, map)
      \ENDWHILE
 \end{algorithmic} 
 \end{algorithm}

\subsection{Initialization}
\label{sec:init}
Initialization occurs in line 1. If initial positions are unknown, TEAM can determine initial positions given that each robot has a known and unique identifier. Robot 0's position defines the map origin and the forward direction. Robot 1's position defines the y axis and robot 2's position defines the positive direction of the x axis. Robot 3 and any additional robots can then trilaterate as discussed below. If initial orientations are unknown, each robot can move a known distance in its local forward direction and then trilaterate to determine orientation. If initial positions are known, TEAM leverages this data to perform sensor auto-calibration, estimating the sensor offset from the truth and using that bias to correct future measurements.

\subsection{Trilateration}
\label{sec:trilaterate}
\begin{figure}
    \centering
    \includegraphics[width=4cm]{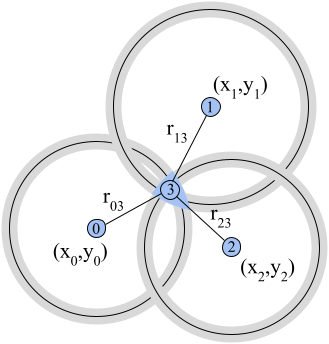}
    \caption{Robot 3 trilateration, using the received position estimates of robots 0, 1, and 2 as anchors.}
    \label{trilateration}
\end{figure}

Robots receive position estimates from their neighbors via RadioReceive in line 3, and collect range measurements via UWBRange in line 6. Given the positions and distances from the three closest instantaneous anchors, TrilateratePosition (line 7) determines a position estimate as shown in Figure \ref{trilateration}. The grey rings represent uncertainty in the two-way ranging estimate, and the estimated position is the centroid of the curved triangle formed by their intersection points. 
If fewer than three anchors are within the communication range, TrilateratePosition relies on the robot's previous position estimate and odometry to determine an updated position estimate. 
For a robotic network of at least five robots, TEAM can trilaterate in three dimensions.

\begin{figure}[ht]
    \centering
    \includegraphics[width=\columnwidth]{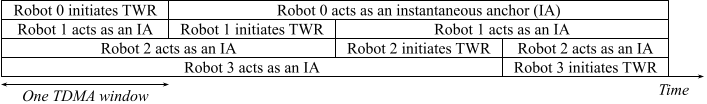}
    \caption{Time-division multiple access scheme for 4 robots performing TEAM. Each robot initializes two-way ranging (TWR) during its TDMA window, and this cycle repeats.}
    \label{fig:TDMAwindsows}
\end{figure}

To prevent UWB signal interference, TEAM uses a time-division multiple access (TDMA) protocol such that lines 5-7 are only executed during a specified window as illustrated in Fig. \ref{fig:TDMAwindsows}. We selected TDMA because it offers high channel utility and has extensions suitable for large teams of cooperating robots \cite{nguyen2016ultra,oliveira2015multi}.

\subsection{Mobility}
\label{sec:drive}
Exploration algorithms which seek to drive the robot toward unexplored areas and maximize new information are presented in \cite{yamauchi1997frontier,yamauchi1998frontier,corah2017efficient,julian2012distributed}, and our previous work \cite{clark2021queue}. TEAM is suitable for both autonomous exploration and teleoperation. In this work we are focused on the resource-constrained setting and elect to use simple drive control (line 5) in which robots are given a target direction and perform autonomous collision avoidance based on LiDAR data.

To improve positioning accuracy, robots are stationary outside of their TDMA window. Note that with four robots, this scheme takes four times as long to cover a sum total distance when compared to all robots driving simultaneously. This highlights an important system design trade-off between prioritizing quick or accurate map coverage. For larger networks, the three robots required to act as instantaneous anchors are a smaller fraction of the network and the reduction in sum total distance covered is less significant.

\subsection{Mapping}
\label{sec:map}
Each time the LiDAR takes a measurement (line 9), this data is associated with the current position estimate. The occupancy grid map representation is then updated (line 10) with the LiDAR data. The resulting map is shared with the data sink via RadioTransmit in line 11. To reduce the burden on the resource-constrained robots, the data sink is responsible for merging the maps received from all sources. Using the map merging algorithm from \cite{horner2016map}, the data sink is able to perform feature mapping and calculate the appropriate frame transforms to merge all received maps. This process does not require synchronization between robots, and the data sink can receive a map as long as any multi-hop communication path is available, as discussed below.







\subsection{Communication}
Under the assumption that all robots are within communication range, the publish/subscribe paradigm provided by the Robot Operating System (ROS) is sufficient for sharing positioning and map data (lines 3 and 11). However, lava tubes are characterized by branches which challenge connectivity \cite{greeley1971lava}. Previous work exploring tunnels has considered a variety of approaches including data tethers and droppable network nodes \cite{tatum}. This work builds on the approach of \cite{cretise} with mobile network nodes which can forward or re-route data. The robots and data sink form an ad hoc mesh network, in which robots can act as relays to the data sink when needed, using Optimized Link State Routing (OLSR) to determine network neighbors and appropriate routes \cite{clausen2003optimized}.

\section{Complexity}
\label{complexity}

In TEAM, selecting the three nearest anchors from the anchors within range can be done in $O(NlogN)$ where $N$ is the size of the network. Calculating the location estimate is then a constant number of operations and several closed-form and approximate algorithms exist \cite{zhou2009efficient}.
Associating scan data with a location estimate and updating the map can also be done in constant time, as new scans are integrated independent of the size of the existing map.

We compare TEAM with an improved Rao-Blackwellized Particle Filter (referred to here as SLAM), a Monte Carlo localization algorithm which is available via the GMapping library \cite{grisetti2007improved, gmapping}.
Following the complexity analysis in \cite{grisetti2007improved}, SLAM introduces complexity $O(P)$ each time the location estimate is updated, where $P$ is the number of particles. This computation is associated with computing the proposal distribution, computing the particle weights, and testing if a resample is required. SLAM also introduces complexity $O(P)$ for each map update, and complexity $O(PM)$ each time a resample occurs, where $M$ is the size of the map. For an optimized system, the number of particles required is typically between 8 and 60 \cite{grisetti2007improved}, depending on the size and features of the environment. Thus, TEAM can provide up to a 60x reduction in computational complexity.
Table \ref{table:runtimes} contains the average runtimes for relevant computations on our hardware.



\begin{table}[ht]
  \centering
  \begin{tabular}{|c c c c|}
    \hline
    Computation & SLAM, P=60 & SLAM, P=1 & TEAM \\
    \hline \hline
     Trilaterate & - & - & 0.001 s \\ 
     \hline
     Update map & 0.40 s & 0.1 s & 0.1 s \\
     \hline
     Process scan & 2.25 s & 0.05 s & $<$ 0.001 s \\
     \hline
  \end{tabular}
  \caption{Runtime comparison}
  \label{table:runtimes}
\end{table}

\section{Simulation Experiments}

\subsection{Simulation Setup}


We use the Gazebo robotics simulator to conduct experiments in various environments. We implemented UWB ranging in simulation according to the following model:
\begin{equation}
    d_{pozyx}=\begin{cases}
    d_{true} + N(\mu=0, \sigma=10cm), \text{ if } 1_{LOS}\\
    \text{None, otherwise}
    \end{cases}
\end{equation}
where $d_{true}$ is the true distance between two robots and $1_{LOS}$ is an indicator function which evaluates to true if and only if line of sight is available.

Table \ref{table:parameters} presents relevant parameters used in the simulation experiments. We assume UWB measurements have zero mean and a standard deviation of 10cm \cite{pozyx}. Each call to UWBRange returns the average of 10 independent measurements. To ensure synchronization between the LiDAR scans and UWB ranging measurements, a delay of more than 100 ms between the two sensors prevents TEAM from updating the map with this scan data. This timeout was empirically selected.

\begin{table}
  \centering
  \begin{tabular}{|l r|}
    \hline
    Parameter & Value \\
    \hline \hline
     Number of robots & 4 \\
     Number of UWB measurements averaged & 10 \\ 
     Std dev of UWB measeurements & 10 cm \cite{pozyx} \\ 
     LiDAR/Trilateration synchronization timeout & 100 ms \\ 
     Map publish rate & 1 Hz \\
     Position estimate publish rate & 10 Hz \\
     Max drive speed & 0.22 m/s \\
     $360^{\circ}$ LiDAR sample rate & 5 Hz \\
     $360^{\circ}$ LiDAR resolution & $1^{\circ}$ \\
     $360^{\circ}$ LiDAR standard deviation & 0.01 \\
     RBPF number of particles & 50 \\
     Odometry variance & 0.01 mm \\ 
     \hline
  \end{tabular}
  \caption{Relevant parameters for simulation experiments}
  \label{table:parameters}
\end{table}

Note that with time-division mobility in the drive module, our approach generates maps at a slower pace. In simulation, we choose to not model the UWB interference constraint and allow the robots to range simultaneously and therefore drive simultaneously. This allows us to compare SLAM and TEAM across the same time scale.



\begin{figure}
    \centering
    \includegraphics[width=\columnwidth]{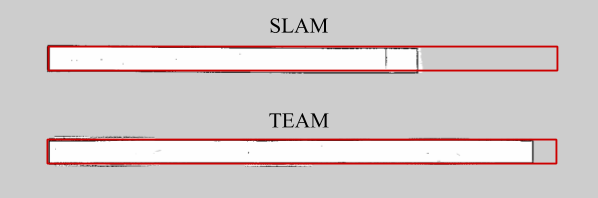}
    \caption{The top image resulted from SLAM with 50 particles, the bottom image resulted from TEAM, each after 25 simulated minutes. The true environment dimensions are overlaid in red.}
    \label{infinite}
\end{figure}

\begin{figure}
    \centering
    \includegraphics[height=5cm]{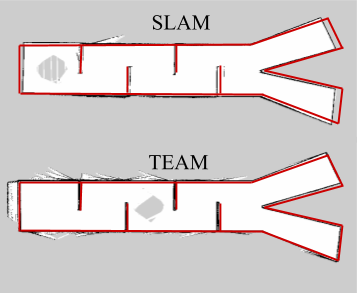}
    \caption{The top image resulted from SLAM with 50 particles, the bottom image resulted from TEAM. The true environment dimensions are overlaid in red.}
    \label{ymaze}
\end{figure}

\begin{figure}
    \centering
    \includegraphics[width=\columnwidth]{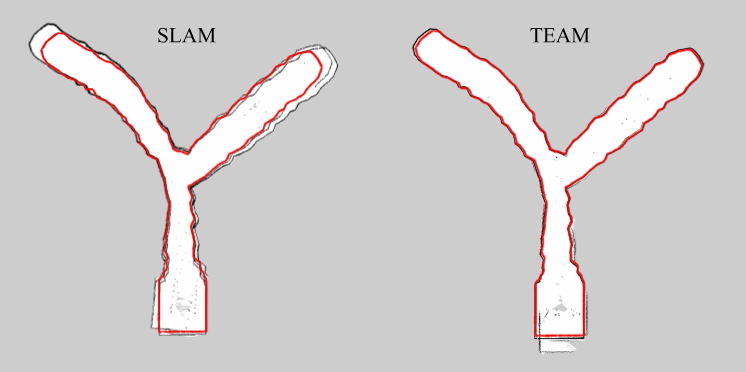}
    \caption{The left image resulted from SLAM with 50 particles, the right image resulted from TEAM. The true environment dimensions are overlaid in red.}
    \label{branches}
\end{figure}

\subsection{Simulation Results}

The first environment we consider challenges SLAM but is well-suited for TEAM: a long, featureless, obstacle-free corridor. This environment is difficult for a particle filter because the particle distribution becomes spread as the robots navigate along the hallway which lacks discernible features for localization \cite{grisetti2007improved}. The results of this experiment are shown in Figure \ref{infinite}; they indicate that TEAM can significantly improve map accuracy in featureless environments.

The second environment we consider challenges TEAM but is well-suited for SLAM: an environment with obstacles and branches that prevent line of sight. TEAM relies heavily on odometry data in this environment, and the results of this experiment are shown in Fig. \ref{ymaze}. We  demonstrate the performance in the presence of noisy odometry data in Sec. \ref{experiments}.

Finally, we consider an environment with lava tube-like characteristics, including two branches. The Gazebo model for this environment comes from \cite{tunnel}. The results of this experiment are shown in Fig. \ref{branches} and strengthen our claim that TEAM results in accurate maps regardless of the environment. 


\begin{figure}
    \centering
    \includegraphics[height=4cm]{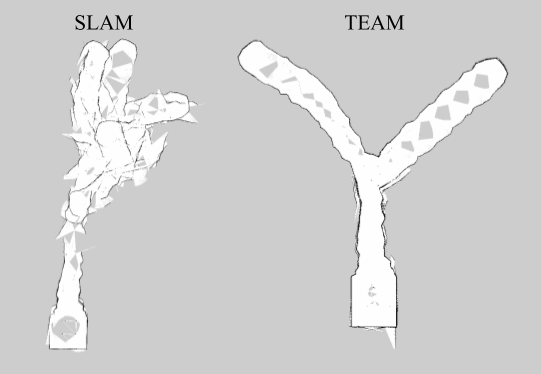}
    \caption{The left image resulted from SLAM with 50 particles and scans throttled to the low frequency of 0.1Hz. The right image resulted from TEAM with the same low LiDAR sample rate.}
    \label{throttle}
\end{figure}

\begin{figure}
    \centering
    \includegraphics[width=\columnwidth]{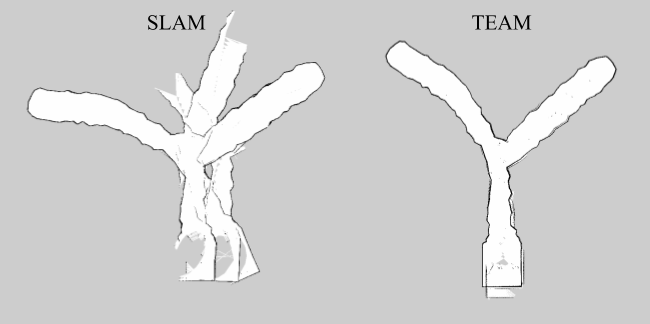}
    \caption{The left image resulted from SLAM with 1 particle. The right image resulted from TEAM, which is computationally comparable.}
    \label{reducedcomputation}
\end{figure}

One significant advantage of TEAM is that it decouples localization from LiDAR sensor measurements. This allows us to reduce the frequency of LiDAR scanning in order to save power; this is useful for robots that have severe energy constraints or memory limitations. Fig. \ref{throttle} shows the result of an experiment in which scans are throttled to 0.1Hz from 5Hz used in the previous experiments. This shows that reduced LiDAR frequency significantly deteriorates the performance of SLAM, while leading to topologically correct albeit sparse maps with TEAM.

As discussed in Sec. \ref{complexity}, the algorithmic complexity of SLAM is a function of the number of particles used to capture the belief distribution of the location estimate. For TEAM, the algorithmic complexity is comparable to SLAM with a single particle. In Fig. \ref{reducedcomputation} we illustrate the performance of SLAM with 1 particle and TEAM side by side, to highlight the difference in map accuracy despite similarly reduced computation time.

\begin{figure}
    \centering
    \includegraphics[width=\columnwidth]{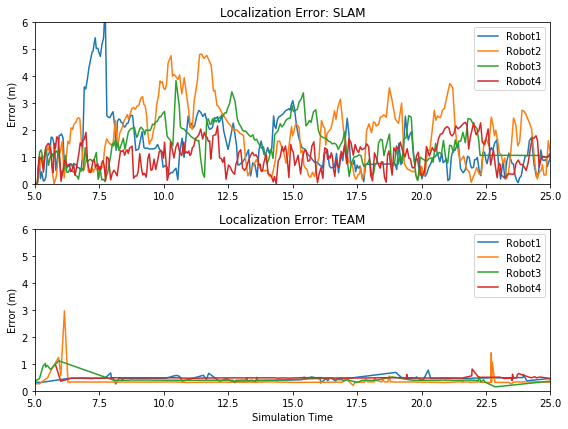}
    \caption{These graphs show the localization error over time for SLAM and TEAM for the environment depicted in Fig. \ref{branches}.}
    \label{loc_error}
\end{figure}

\begin{figure}
    \centering
    \includegraphics[height=3.5cm]{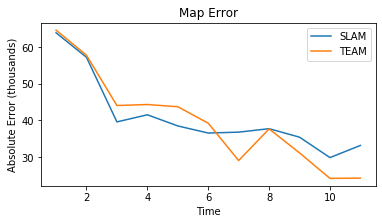}
    \caption{This graph shows the map error as a function of time for the environment depicted in Fig. \ref{branches}.}
    \label{map_error}
\end{figure}

While the merged map presents a clear picture of the capabilities of TEAM, it is also useful to consider the localization error of the robots over time. For SLAM, the localization uncertainty varies as the belief distribution changes over time. In feature-deprived environments like the one shown in Fig. \ref{infinite}, this error can grow up to 25m. Fig. \ref{loc_error} compares the localization error as a function of time during exploration of the lava tube-like environment. We observe that errors in TEAM are less than 3m and these errors are infrequent, however errors in SLAM can grow up to 6m and are higher on average.

It is similarly useful to quantify the accuracy of the merged map created by each algorithm over time. We measure the absolute pixel error of map image files which were manually aligned with and compared to the image file of the true environment. We plot the map accuracy over time and show that while SLAM and TEAM both result in decreasing map error, TEAM achieves a lower final absolute pixel error (Fig. \ref{map_error}). We observe that TEAM achieves a 28\% decrease in error relative to SLAM for the environment depicted in Fig. \ref{infinite}, and a 27\% decrease in error for the environment depicted in Fig. \ref{branches}.






\section{Testbed Experiments}

\begin{table}
  \centering
  \begin{tabular}{|l r|}
    \hline
    Parameter & Value \\
    \hline \hline
     TDMA window size & 5 s \\
     UWB ranging rate & 20 Hz \\
     Worst case expected communication delay & 400 ns \\
     \hline
  \end{tabular}
  \caption{Additional relevant parameters for testbed experiments}
  \label{table:additionalparameters}
\end{table}

\subsection{Testbed Setup}

\begin{figure}
    \centering
    \includegraphics[width=0.45\columnwidth,trim={300 150 200 300},clip,angle=270]{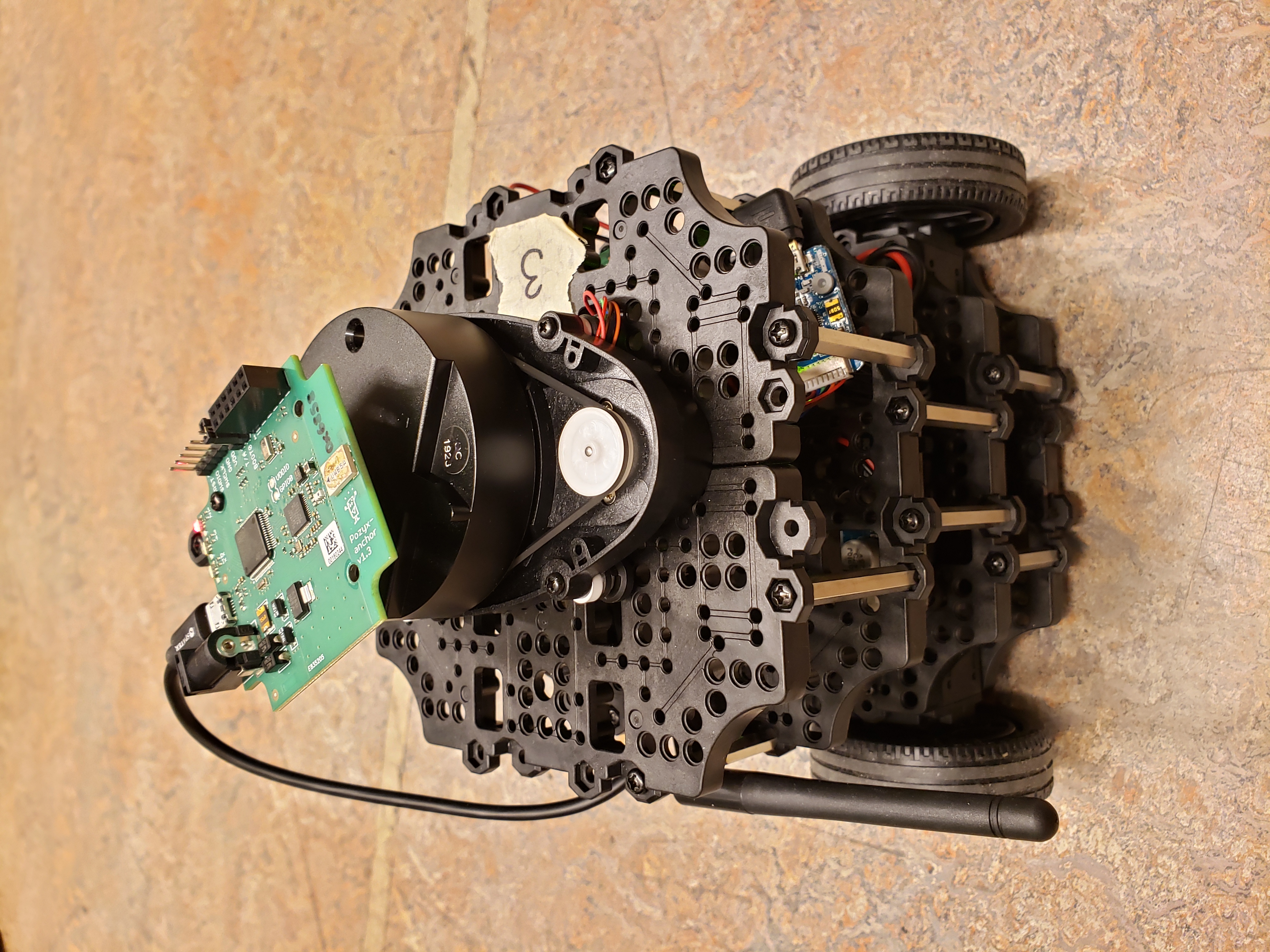}
    \caption{Turtlebot3 Burger with Pozyx UWB anchor.}
    \label{turtlebot3}
\end{figure}

Our testbed is comprised of four Turtlebot3 Burgers, each equipped with a RaspberryPi 3B+ running Raspbian, an OpenCR1.0 control board, a 360 Laser Distance Sensor, and differential drive. We have extended each platform with a Pozyx UWB Creator series Anchor, and a USB Wireless Adapter Mideatek RT5370N with 2dBi antenna. The wireless adaptor allows each robot to designate one wireless interface for internet connectivity, and one wireless interface for joining our ad hoc mesh network, TurtleNet. The robots and data sink, a PC running Ubuntu 16.04, all implement Optimized Link State Routing (OLSR), a proactive routing protocol for mobile ad hoc mesh networks \cite{clausen2003optimized, tonnesen2008olsrd}. We provide a tutorial for mesh networking Turtlebot3 Burgers \cite{olsr_repo}.

Table \ref{table:additionalparameters} presents parameters used in the testbed experiments in addition to those previously listed in Table \ref{table:parameters}. The size of the TDMA window should be larger than the time to collect 10 ranging measurements (0.5 sec) and smaller than the time it would take a robot to drive beyond the maximum communication range (90 sec). Delays between the robots and the data sink do not affect the quality of the resulting merged map, as it is processed asynchronously; however, delays or dropped packets between robots affect the location estimate accuracy. We introduce a 0.3 second buffer period at the end of each TDMA window during which the robot stops driving but continues to perform trilateration. Note that for consistency, our time-division approach to driving autonomously was used across all experiments.

\begin{figure}[ht]
    \centering
    \includegraphics[height=3cm]{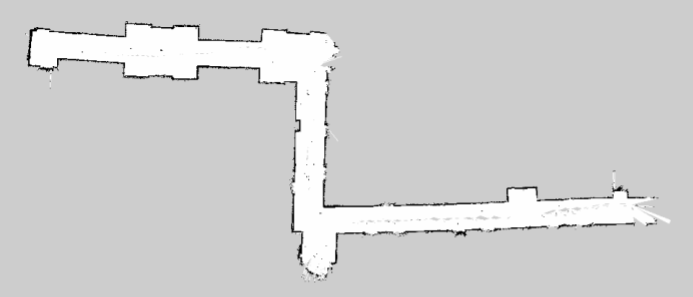}
    \caption{SLAM in the hallway: merged map}
    \label{slam_composite}
\end{figure}

\begin{figure}[ht]
    \centering
    \includegraphics[height=3cm]{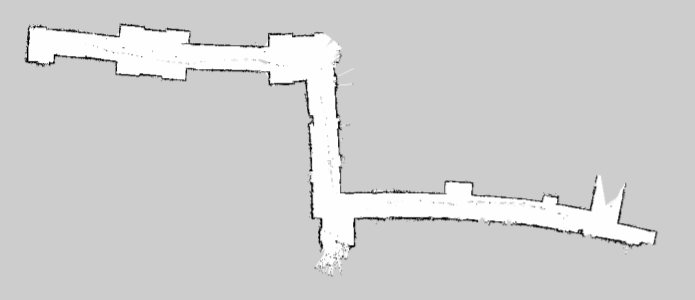}
    \caption{SLAM in the hallway: map created by a single robot}
    \label{slam_separate}
\end{figure}

\begin{figure}[ht]
    \centering
    \includegraphics[height=3cm]{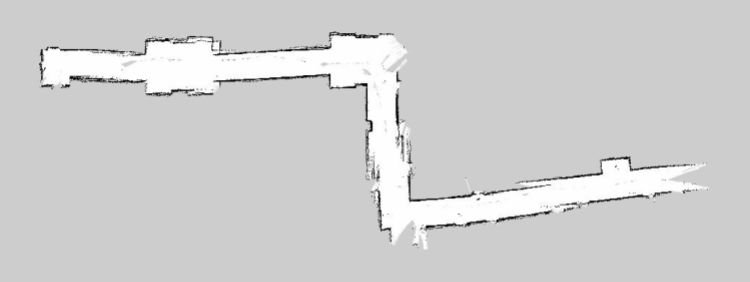}
    \caption{TEAM in the hallway: merged map}
    \label{team_composite}
\end{figure}

\begin{figure}[ht]
    \centering
    \includegraphics[height=3cm]{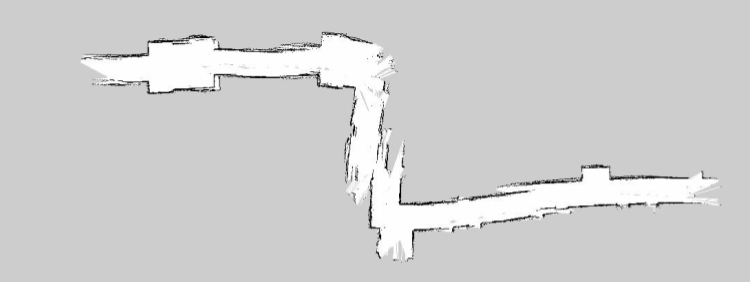}
    \caption{TEAM in the hallway: map created by a single robot}
    \label{team_separate}
\end{figure}

\subsection{Testbed Results}
\label{experiments}

Fig. \ref{slam_composite} shows the merged map created by our testbed in a typical hallway environment using SLAM. Fig. \ref{slam_separate} shows the map created by a single robot, illustrating the effect of inaccuracy in the odometry readings. We noticed that the drift in odometry was not consistent across the four robots, but the map merging algorithm mitigated this. For better location estimate accuracy, the robots should localize within the merged map rather than their individual maps.

Fig. \ref{team_composite} shows the merged map created using TEAM. We observe that the wall edges are slightly less well-defined. This is due to noise in the UWB ranging measurements causing small jumps in the position estimate. This noise is primarily attributed to multipath effects, and further characterizing the UWB performance will help improve map accuracy \cite{prorok2014accurate}. Fig. \ref{team_separate} shows the individual map of a single robot and highlights the effect of loss of line of sight. This causes the robot to rely on its odometry until UWB signals are available again, at which point its location estimate may jump, leading to discontinuities in the map. Distributed strategies to maintain connectivity as presented in \cite{clark2021queue} can prevent these discontinuities.










\section{Conclusions and Future Work}

We have presented the design, implementation, and evaluation of TEAM, a novel algorithm for localization and mapping in unknown environments with a robotic network. We demonstrate the ability of TEAM to leverage ultra-wideband positioning to generate maps of various environments with high accuracy. Our algorithm significantly reduces the computational complexity and the required rate of LiDAR samples, making it suitable for resource-constrained multi-robot systems, and performs well in feature-deprived environments where SLAM struggles.

In the future, we will investigate the use of discrete or continuous belief distributions over the location estimate to better capture the randomness in the Pozyx signals. We also plan to further characterize the UWB multi-path effects. Other directions for future research could include incorporating a consensus mechanism for electing instantaneous anchors \cite{aragues2012distributed, tian2008consensus}, or improving inter-robot synchronization \cite{sheu2007clock, cao2004accurate}.

\bibliographystyle{IEEEtran}
\bibliography{bibliography}

\end{document}